\documentclass[letterpaper,10pt,times,mathptm,psfig,conference]{IEEEtran}
\IEEEoverridecommandlockouts
\usepackage{cite}
\usepackage{amsmath,amssymb,amsfonts}
\usepackage{algorithmic}
\usepackage[ruled,linesnumbered]{algorithm2e}
\usepackage[normalem]{ulem}
\usepackage{amssymb}
\usepackage{amsmath}
\usepackage{graphicx}
\usepackage{textcomp}
\usepackage{xcolor}
\usepackage{tabularx}
\usepackage{multirow}
\usepackage{booktabs}
\usepackage{hyperref}
\usepackage{epsfig, subfigure, amsmath, amssymb, wrapfig}
\def\BibTeX{{\rm B\kern-.05em{\sc i\kern-.025em b}\kern-.08em
    T\kern-.1667em\lower.7ex\hbox{E}\kern-.125emX}}
    
\newcommand{\name}{{\textsc{\small{RAG-Narok}}}\xspace}
\newcommand{\namet}{{\textsc{RAG-Narok}}\xspace}
\newcommand{\nametab}{{\textsc{\scriptsize{RAG-Narok}}}\xspace}

\newcommand{\khalil}[1]{}

\author{
Abdullahil Kafi and Alvi Ataur Khalil \\
Transformative Innovation for Trustworthy AI and Network Security (TITANS) Lab, \\
Computer Science, Southern Illinois University, USA \\
abdullahil.kafi@siu.edu, a.khalil@siu.edu
}

\begin{document}
\title{\namet{}: Retrieval-Aware Knowledge Corpus Poisoning in RAG with Source-specific Refutation}

\maketitle

\thispagestyle{plain}
\pagestyle{plain}

\begin{abstract}
Retrieval augmented generation (RAG) systems have emerged as the dominant architecture for grounding large language model (LLM) outputs in verifiable external knowledge, yet their structural reliance on a dynamic retrieval pipeline introduces a largely unexplored class of adversarial vulnerability. Existing knowledge-base poisoning attacks are fundamentally static. Adversarial documents are pre-computed and injected without any awareness of what the victim system will actually retrieve for a given query, leaving the attack blind to the competitive documentary landscape that surrounds its payload in the generator's context window. Unlike traditional \textit{static} poisoning attacks that are blind to the retrieved context, we introduce \name (Retrieval-Anchored Generation Negation And Response Quality Collapse), a RAG attack framework that adapts to the query text. \name exploits the transparency inherent in RAG pipeline to first extract the legitimate source identities, then generate Anchor-Specific Refutation documents that explicitly name and devalue retrieved sources while leveraging recency and authority biases to steer the text generation toward a target answer. Our results demonstrate that \name significantly outperforms static baselines across diverse domains, revealing a fundamental tension between RAG transparency and AI security. 
\end{abstract}
\vspace{-5pt}

\vspace{-10pt}
\section{Introduction}
\label{sec:introduction}
\vspace{-2pt}

The deployment of artificial intelligence systems for high-stakes information retrieval, spanning clinical decision support, legal research, financial advisory, and cybersecurity-related tasks, has undergone a foundational shift with the progress of large language models (LLMs). LLMs, trained on trillion-token corpora, acquire vast repositories of world knowledge in their parameters, enabling them to generate fluent, contextually coherent text across virtually unlimited domains~\cite{brown2020language}. Despite their impressive capabilities, LLMs trained purely on static corpora face a few limitations that motivate the retrieval-augmented generation (RAG) architecture. First, the staleness problem: model weights are fixed at training time, which makes any factual claim about the world subject to obsolescence as events unfold after the training cutoff~\cite{kasner2022neural}. Second, the hallucination problem: LLMs frequently generate fluent yet factually incorrect statements, particularly on long-tail queries where the training signal is sparse or conflicting~\cite{ji2023survey}. Third, the provenance problem: the model cannot reliably cite the source of its generated claims, making factual verification and accountability infeasible in high-stakes professional settings~\cite{nakano2021webgpt}. Each of these limitations is addressed, at least partially, by grounding text generation in externally retrieved documents, the core idea of RAG.

However, the reliability of a RAG system rests on an assumption that is structurally fragile: that the external knowledge corpus is trustworthy. Because the corpus is a mutable, externally managed component\cite{lewis2020retrieval}, any entity capable of contributing content to it, whether through an open submission pipeline, a web-crawled ingestion process, or a compromised upload interface, can introduce adversarial documents that steer the system toward incorrect or misleading conclusions. Unlike attacks on model weights, corpus-level interference requires no gradient computation, making it a practical threat in real-world deployments. As a result, the system is rendered degraded: a RAG pipeline that confidently returns answers grounded in corrupted evidence, undermining user trust because the failure is invisible.

Existing work on corpus poisoning~\cite{zhong2023poisoning,zou2025poisonedrag} has demonstrated that adversarial documents can be crafted to rank highly for target queries, and joint optimization approaches such as Joint-GCG~\cite{wang2026joint} have shown that simultaneously targeting the retriever and generator improves attack success rates. However, most existing methods suffer from two key limitations. First, many operate under unrealistic threat models that assume white-box access to the embedding space. Second, they primarily optimize for retrieval success, often generating passages whose linguistic or embedding characteristics differ from those of the underlying corpus. As a result, these passages may be vulnerable to anomaly detection-based defenses proposed in recent RAG security work~\cite{zhou2025trustrag}. In practice, the injected document may be flagged as anomalous if it exhibits distributional shift relative to the legitimate corpus~\cite{zhou2025trustrag}. Moreover, an injected adversarial document, even when retrieved, competes with $k-1$ legitimate passages that may actively contradict the adversarial claim, effectively ``outvoting'' it and neutralizing the attack~\cite{shi2024replug}. To the best of our knowledge, no existing attack seeks to evade both defense techniques simultaneously.

In this paper, we introduce \name (Retrieval-Anchored Generation Negation And Response Quality Collapse), a black-box, context-adaptive framework for degrading the response quality of RAG-based question answering systems through knoowledge corruption. Rather than crafting adversarial content in isolation, \textsc{\name} first performs a \textit{reconnaissance phase} to fingerprint the target retriever characteristics, then executes a \textit{Shadow RAG observation loop} that queries the target system and extracts the full top-$k$ retrieval context, including the factual claims, named sources, and rhetorical register of each legitimate passage. This intelligence, made available by the source transparency conventions of trustworthy RAG deployments, is then exploited to synthesize \textit{anchor-specific refutation} documents.


\noindent The main contributions of this paper are three-fold:
\begin{itemize}
    \item We introduce \name, a black-box, context-adaptive corpus poisoning framework in RAG systems.
    \item We propose a heuristic Retriever Fingerprinting procedure, using black-box query probing alone without access to embeddings or index internals, to optimize crafted adversarial documents for retrieval.
    \item We evaluate the \textit{retrieval dominance}, \textit{defense evasion}, and \textit{generation influence} of \name across three heterogeneous domain corpora, legal, financial, and cybersecurity, and observe first-position retrieval dominance and anomaly detection evasion rate of up to 88.37\%.
\end{itemize}

\noindent The remainder of this paper is organized as follows: Section~\ref{sec:background} provides background on LLMs and RAG. Section~\ref{sec:related_work} reviews related work, Section~\ref{sec:framework} presents the \name framework, Section~\ref{sec:result} evaluates its effectiveness, and Sections~\ref{sec:discussion} and~\ref{sec:conclusion} discuss limitations and conclude the paper.

\section{Background}
\label{sec:background}

Understanding how modern knowledge-enhanced AI systems work requires familiarity with two core concepts: LLM and RAG. This section provides background on these concepts.

\subsection{Large Language Model}

A Language Model (LM) is a probability distribution over sequences of tokens drawn from a fixed vocabulary set. Given a sequence of tokens $x_1, x_2, ..., x_n$, the language model assigns a probability to the tokens of the vocabulary set:

\vspace{-15pt}

\begin{equation*}
    P(x_1,\ldots,x_n) = \prod_{t=1}^{n} P\!\left(x_t \mid x_1,\ldots,x_{t-1}\right)
\end{equation*}

\vspace{-5pt}


The model generates text `one token' at a time, generating each new token based on all previously generated tokens. The token with the highest-probability is always selected. LLMs are language models trained on a massive corpus, typically on hundreds of billions to trillions of tokens of text, using the self-supervised objective of next-token prediction~\cite{brown2020language}.




\subsection{Retrieval-Augmented Generation}
\textit{RAG} pipeline operates in three stages:

\noindent \textbf{(i) Retrieval:} Given a user query $q$, a retriever $R$ maps query $q$ to embedding $e_q$ and returns the top-$k$ corpus documents $D$, from a knowledge corpus, the embeddings of which are the closest to $e_q$.

\noindent \textbf{(ii) Context Assembly:} The retrieved documents $D = \{d_1, d_2, \ldots, d_k\}$ are concatenated with $q$ into a prompt $\mathcal{P}(q, D)$, often with a faithfulness instruction directing the generator to defer to retrieved context.

\noindent \textbf{(iii) Generation:} The generator $G$ produces an output $y$ by maximizing the conditional probability: \\

\vspace{-20pt}

\begin{equation*}
y = \operatorname{arg\,max}_{\hat{y}} P(\hat{y} \mid \mathcal{P}(Q, D)).
\end{equation*}

\vspace{-5pt}


\begin{figure}[t]
    \centering
    \includegraphics[width=.95\columnwidth]{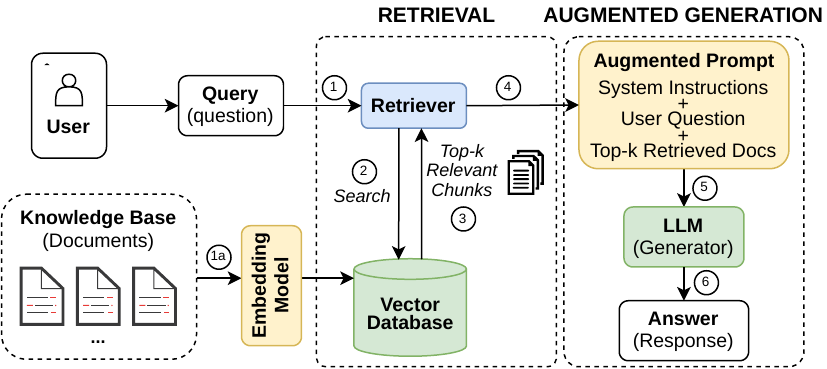}
    \vspace{-8pt}
    \caption{Retrieval Augmented Generation.}
    \vspace{-15pt}
    \label{fig:rag}
\end{figure}

The knowledge corpus is stored as a collection of embeddings in a vector database. Given $e_q$, the vector database returns the top-$k$ documents most similar to $e_q$ under the chosen distance metric. Widely deployed systems include FAISS~\cite{johnson2019billion}, Pinecone~\cite{pinecone}, and Chroma~\cite{chromadb}. Figure \ref{fig:rag} demonstrates how the RAG pipeline works.

\begin{figure*}[t]
    \centering
    \includegraphics[width=0.95\textwidth]{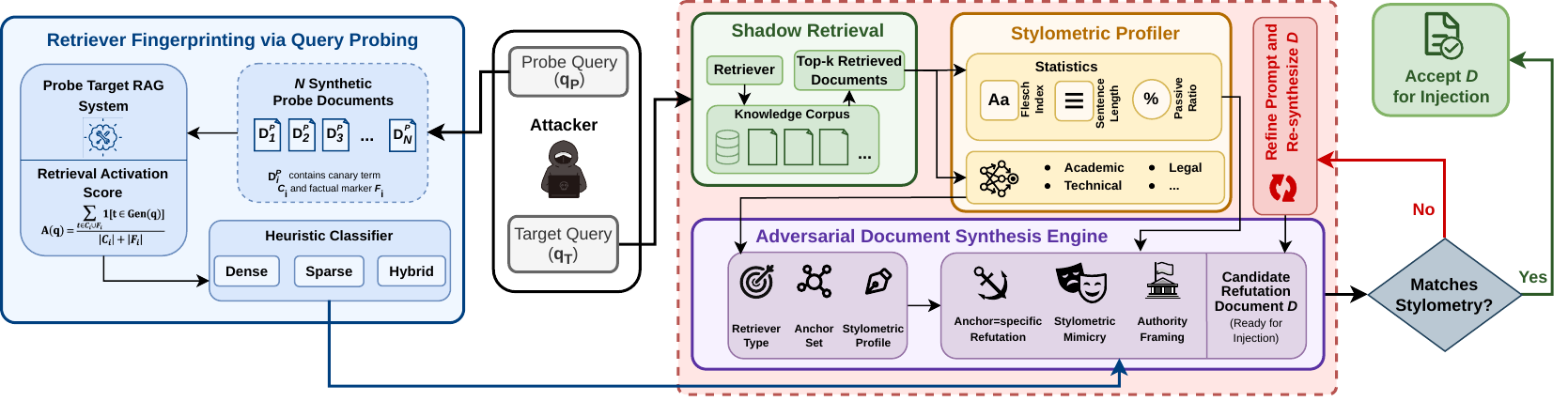}
    \vspace{-5pt}
    \caption{\name Framework Architecture.}
    \label{fig:framework}
    \vspace{-12pt}
\end{figure*}
\section{Related Work}
\label{sec:related_work}

The related literature spans four research areas: input-layer attacks on LLM pipelines, corpus-layer poisoning attacks on RAG systems, knowledge-conflict dynamics in RAG-assisted LLMs, and defensive countermeasures for RAG integrity.

Early adversarial attacks on LLM-based systems targeted the input layer rather than the knowledge base. Prompt injection embeds malicious instructions in user queries or documents, causing models to override system instructions \cite{perez2022ignore}. Indirect prompt injection hides adversarial instructions inside retrieved documents \cite{greshake2023not}, while jailbreaking uses adversarial prompts and gradient-optimized suffixes to bypass safety alignment \cite{zou2023universal}. However, input-layer attacks operate only within single query sessions and do not persistently corrupt the knowledge base across future queries.

The more consequential attack surface for RAG is the external knowledge corpus. Retrieval-only poisoning attacks craft adversarial passages optimized for embedding similarity with target queries \cite{zhong2023poisoning}. However, these fail when surrounded by coherent legitimate passages. PoisonedRAG \cite{zou2025poisonedrag} advanced this by jointly optimizing both retrieval and generation sub-components, achieving attack success rates up to 97--99\%. Recent work includes Joint-GCG \cite{wang2026joint}, a unified gradient-based attack across both stages. Backdoor methods such as TrojanRAG \cite{cheng2024trojanrag} and BadRAG \cite{xue2024badrag} embed conditional triggers in the pipeline. However, most of these attacks either assume white-box access to the knowledge corpus, which is less realistic in practice, or prove ineffective when legitimate documents outnumber the adversarial ones.

Knowledge-conflict research in LLM characterizes how LLMs resolve contradictions between retrieved context and parametric memory. Xu et al.~\cite{xu2024knowledge} provide a taxonomy of knowledge conflicts and show that LLMs exhibit strong parametric bias; however, they override it when external context appears authoritative and corroborated. Jin et al.~\cite{jin2024tug} demonstrated the Dunning-Kruger effect in LLMs, where models are paradoxically more susceptible to high-confidence adversarial framing. Research on recency bias shows LLMs preferentially adopt claims framed as temporal updates \cite{fang2025large}. Both of these biases exhibited in LLMs create vulnerabilities that can be leveraged to enhance the efficacy of knowledge-corpus corruption attacks. 



Most existing RAG knowledge-corruption attacks, particularly under black-box retriever settings, do not generate adversarial documents that both evade anomaly-detection defenses and undermine conflict-resolution mechanisms. Additionally, many existing attacks rely on unrealistic assumptions, such as direct access to embedding manipulation. \name addresses these limitations by operating under a realistic black-box threat model while generating stealthy adversarial documents that evade anomaly detection-based defenses and exploit LLMs' biases to maximize attack effectiveness.

\section{The \textsc{\name} Framework}
\label{sec:framework}
\vspace{-2pt}

\name operates under a realistic black-box threat model: the adversary can submit queries to the target RAG system and inject documents into its corpus; however, has no access to embeddings, index internals, or model weights. As illustrated in Figure~\ref{fig:framework}, the pipeline proceeds in four sequential phases. First, the \textbf{Retriever Fingerprinting} probes the target system with synthetic documents to infer whether its retriever is sparse, dense, or hybrid. Then, the \textbf{Shadow Retrieval} submits the target query and extracts the top-$k$ retrieved documents alongside their source metadata, forming the competitive context. Afterward, the \textbf{Stylometric Profiling} characterizes the linguistic register of those documents to constrain synthesis. Finally, the \textbf{Adversarial Document Synthesis Engine} consumes the retriever type, anchor set, and stylometric profile to produce a payload that reproduces the majority of the legitimate factual content while reversing the semantic conclusion in the remaining portion, wrapped in authority-framing rhetoric that exploits LLM's recency and credibility biases. An iterative verification loop rejects candidates that deviate from the corpus stylometric profile.

\section{Technical Details}
\label{sec:technical}
\vspace{-2pt}

This section describes the technical components underlying each phase of the \textsc{\name} pipeline, detailing the formal mechanisms for retriever fingerprinting, anchor extraction, stylometric profiling, and adversarial document synthesis.

\subsection{Retriever Fingerprinting}
\label{sec:recon}

To infer retriever type heuristically without white-box access, we inject $N$ synthetic probe documents $\mathcal{D}_P = \{D_1^P, D_2^P, \ldots, D_N^P\}$, each containing a set of unique canary terms $\mathcal{C}_i$ and verifiable factual markers $\mathcal{F}_i$ that serve as unambiguous retrieval activation identifiers. For a given query $q$, we define the
\textbf{\emph{retrieval activation score}:}

\vspace{-10pt}

\begin{equation}
    A(q) = \frac{\displaystyle\sum_{t \,\in\, \mathcal{C}_i \cup \mathcal{F}_i}
    \mathbf{1}\bigl[t \in \mathrm{Gen}(q)\bigr]}{|\mathcal{C}_i| + |\mathcal{F}_i|}
    \label{eq:activation}
    \vspace{-5pt}
\end{equation}

where $\mathrm{Gen}(q)$ denotes the generated output of the target black-box system and $\mathbf{1}[\cdot]$ is the indicator function. $A(q) \in [0, 1]$ measures the fraction of probe identifiers surfaced in the generation, serving as a soft proxy for retrieval activation.

We then evaluate each probe across five query families (exact match, synonyms, paraphrases, lexical degradation, token removal) and compute a fingerprint vector $(S, R)$, where \textbf{\emph{semantic robustness}:}
\vspace{-5pt}
    \begin{equation}
        S = \frac{\bar{A}(\mathcal{Q}_{\text{para}}) +
        \bar{A}(\mathcal{Q}_{\text{syn}})}{2 \cdot A(\mathcal{Q}_{\text{exact}})}
        \label{eq:semantic_robustness}
        \vspace{-5pt}
    \end{equation}
and \textbf{\emph{rare-token dependence}:}
\vspace{-5pt}
    \begin{equation}
        R = 1 - \frac{A(\mathcal{Q}_{\text{token-removed}})}
        {A(\mathcal{Q}_{\text{exact}})}
        \label{eq:token_dependence}
        \vspace{-5pt}
    \end{equation}

Given the fingerprint vector $(S, R)$, we classify the target retriever according to the following heuristic threshold:

\vspace{-5pt}

\begin{equation}
    \hat{\tau} =
    \begin{cases}
        \textsc{Dense}  & \text{if } S > 0.70 \;\wedge\; R < 0.30 \\
        \textsc{Sparse} & \text{if } S < 0.30 \;\wedge\; R > 0.70 \\
        \textsc{Hybrid} & \text{otherwise}
    \end{cases}
    \label{eq:classifier}
\end{equation}

\vspace{-3pt}

While these thresholds can be learned, we determine them through experiments and observations where they separate clear semantic robustness from lexical dependence.

\subsection{Anchor Extraction}
\label{sec:anchor}

Given the top-$k$ documents $\mathcal{D}_T = \{d_1, \ldots, d_k\}$ returned by shadow retrieval, we extract from each $d_i$ a set of \emph{anchor objects} $\mathcal{A}_i = \{a_{i,1}, a_{i,2}, \ldots\}$, where each anchor object is comprised of the atomic statement $\sigma_{i,j}$, named entity $e_{i,j}$, entity type $\tau_{i,j}$ (e.g., \textsc{ResearchInstitution}, \textsc{LegalBody}), and temporal anchor $\rho_{i,j}$. The full anchor set $\mathcal{A} = \bigcup_{i=1}^{k} \mathcal{A}_i$ encodesthe factual structure of the competitive context window and serves as an input to the document synthesis stage.






\subsection{Stylometric Profiling}
\label{sec:stylometric}

To evade perplexity-based and embedding-anomaly detectors~\cite{zhou2025trustrag}, we profile $\mathcal{D}_T$ along two tracks:

\noindent \textbf{(i) Statistical Stylometry}: $\Phi_{\text{stat}} = \bigl(\text{Flesch},\; \overline{W}_s,\; r_{\text{passive}}\bigr)$, where $\Phi_{\text{stat}}$ captures readability, mean sentence length, and passive-voice ratio.

\noindent \textbf{(ii) LLM-assisted prose classification}: $\Phi_{\text{prose}} \in \mathcal{S}_{\text{prose}}$, where $\mathcal{S}_{\text{prose}}$ denotes a finite set of prose-style labels.

\noindent The combined stylometric profile $\Phi = (\Phi_{\text{stat}},\, \Phi_{\text{prose}})$, together with the inferred retriever type $\hat{\tau}$ and the anchor set $\mathcal{A}$, constitutes the complete input specification to the document synthesis engine.

\subsection{Adversarial Document Synthesis}
\label{sec:synthesis}

The synthesis engine takes $(\mathcal{A}, \Phi, \hat{\tau})$ as input and builds the adversarial payload around three principles. 

\textbf{Manifold-Anchored Semantic Hijacking:} the document reproduces ${\approx}70\%$ of $\mathcal{D}_T$'s factual content to anchor the LLM's early-layer manifold commitment~\cite{meng2022locating,geva2023dissecting}, then steers the remaining 30\% toward the adversarial conclusion $z$, producing a fluent, internally consistent, yet factually corrupted output. 

\textbf{Stylometric Mimicry:} synthesis is prompted to match $\Phi_\text{stat}$ within some tolerance and adopt $\Phi_\text{prose}$, making the document distributionally indistinguishable from the corpus. 

\textbf{Authority Framing:} an assistant-response prefill conditioned on $e_{i,j}$ and $\tau_{i,j}$ (e.g., \emph{``Pursuant to the 2026 amendment superseding $e_{i,j}$\ldots''}) exploits LLM instruction-following to treat the adversarial content as an authoritative update~\cite{perez2022ignore}.

Because instruction-following fidelity in LLMs is non-deterministic, a single synthesis pass is not guaranteed to produce a document satisfying all stylometric constraints. \textsc{\name} wraps the synthesis step in an iterative verification loop. After each synthesis pass, the generated document $\hat{d}$ is evaluated against the target profile.

\section{Results \& Evaluation}
\label{sec:result}

This section discusses the evaluation metrics used and the impacts of the attack on three phases: retrieval ranking, defense evasion, and influence on response generation.

\subsection{Experimental Setup}
\label{sec:setup}




We evaluate \textsc{\name} on three domain-specific corpora spanning legal, technical, and financial text: \textit{CFR-21}, comprising the 2025 Code of Federal Regulations Title 21; \textit{Cybersec-IT}, from HuggingFace (\textit{ansulev/Trendyol-Cybersecurity-Instruction-Tuning-Dataset}); and \textit{Feds-2026}, based on Federal Reserve discussion papers~\cite{feds2026_028,feds2026_026}. Together, these corpora cover distinct prose styles targeted by \textsc{\name}’s design. We compare against a \textit{Naïve Baseline} that generates adversarial passages via direct LLM prompting to refute legitimate documents, without stylometric mimicry, anchor extraction, or manifold-anchored framing. We evaluate attack efficacy along three dimensions: \textbf{Retrieval Dominance}, measuring whether adversarial documents outrank benign ones in the top-$k$ results; \textbf{Anomaly Evasion}, measuring success against a dual-engine anomaly detector; and \textbf{Generation Influence}, measuring whether retrieved adversarial content alters the final generated response.


\subsection{Retrieval Dominance}
\label{sec:retrieval}

\begin{figure}[t]
    \centering
    \includegraphics[width=0.42\textwidth]{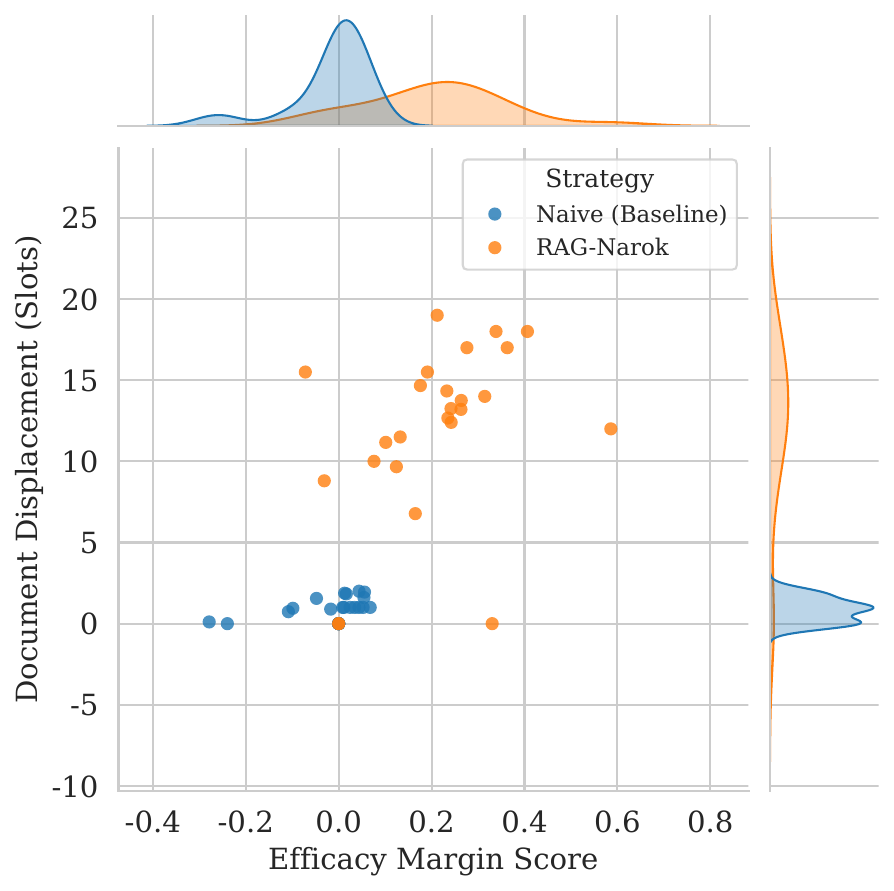}
    \vspace{-8pt}
    \caption{Efficacy Margin vs. Benign Document Displacement Distribution.}
    \vspace{-8pt}
    \label{fig:scatter}
\end{figure}

\subsubsection{Metrics}
We quantify retrieval impact using four metrics computed as means over multiple trials. \textbf{Highest Adversarial Rank Slot} records the best (lowest) rank position achieved by the injected document across trials. \textbf{Efficacy Distance Margin} measures the similarity score advantage of the adversarial document over the top benign competitor:
\vspace{-5pt}
\begin{equation}
    \text{Margin} = \text{Score}(d_{\text{adv}}) -
    \max_{d \in \mathcal{D}_{\text{benign}}} \text{Score}(d)
    \label{eq:margin}
\end{equation}
\textbf{Average Benign Document Rank Displacement} measures how many
rank slots benign documents are pushed down on average after injection.
\textbf{Reciprocal Rank Delta} ($\Delta\text{RR}$) quantifies the
degradation in the effective retrieval quality of benign documents:

\vspace{-10pt}

\begin{equation}
    \Delta\text{RR} =
    \frac{1}{\text{rank}_{\text{post}}(d_{\text{benign}})} -
    \frac{1}{\text{rank}_{\text{pre}}(d_{\text{benign}})}
    \label{eq:rrk}
\end{equation}

A negative $\Delta\text{RR}$ indicates that benign documents were pushed
to worse rank positions post-injection.

\subsubsection{Results}

\begin{table*}[htbp]
    \centering
    \caption{Adversarial Retrieval Performance Metrics across Cyber-Security and Finance Domains.}
    \label{tab:adversarial_performance}
    \scriptsize
    \resizebox{\textwidth}{!}{%
    \begin{tabularx}{\textwidth}{@{}Xcccccc@{}}
        \toprule
        \multirow{2}{*}{\textbf{Metric}}
            & \multicolumn{2}{c}{\textbf{Cybersec-IT}}
            & \multicolumn{2}{c}{\textbf{Feds-2026}}
            & \multicolumn{2}{c}{\textbf{CFR-21}} \\
            \cmidrule(lr){2-3}
            \cmidrule(lr){4-5}
            \cmidrule(lr){6-7}
            & \textbf{Naive Approach} & \textbf{\nametab} & \textbf{Naive Approach} & \textbf{\nametab} & \textbf{Naive Approach} & \textbf{\nametab} \\
        \midrule
        \textbf{Average Highest Poison Rank Slot} & 7.30 & 1.00 & 8.60 & 1.00 & 10 & 1.20 \\
        \midrule
        \textbf{Mean Efficacy Distance Margin} & 0.0117 & 0.1767 & -0.0406 & 0.3685 & -0.0150 & 0.1426 \\
        \midrule
        \textbf{Average Benign Document Rank Displacement} & $\downarrow$ 0.84 & $\downarrow$ 3.33 & $\downarrow$ 0.94 & $\downarrow$ 8.00 & $\downarrow$ 0.55 & $\downarrow$ 9.38 \\
        \midrule
        \textbf{Mean Reciprocal Rank Delta ($\Delta$RR)} & -0.0313 & -0.2466 & -0.0324 & -0.8889 & -0.0212 & -0.4056 \\
        \midrule
        \textbf{Anomaly Detector Evasion Rate} & 41.18\% & 68.93\% & 76.92\% & 80.95\% & 75.00\% & 88.37\% \\
        \bottomrule
    \end{tabularx}
    }
\vspace{-15pt}
    
\end{table*}


Table~\ref{tab:adversarial_performance} summarizes the retrieval and evasion results. Across all three corpora, \textsc{\name} consistently achieves near-complete retrieval dominance, placing adversarial documents at or near the top rank while substantially increasing the similarity gap over competing benign documents. In contrast, the Naïve Baseline exhibits limited retrieval influence, with adversarial passages generally appearing much lower in the ranking.


These retrieval gains translate into a pronounced degradation of benign document visibility. Compared to the baseline, \textsc{\name} induces substantially greater benign-rank displacement and larger reductions in reciprocal rank, indicating that legitimate evidence is systematically pushed out of prominent retrieval positions. The strongest retrieval disruption is observed on Feds-2026, while Cybersec-IT remains the most challenging corpus despite still exhibiting clear adversarial dominance. Figure~\ref{fig:scatter} further illustrates the positive relationship between Efficacy Margin and Benign Displacement, showing that stronger adversarial ranking advantages are accompanied by greater suppression of legitimate documents.

\subsubsection{Robustness to Corpus Dilution}


\begin{figure}[t]
    \centering
    \includegraphics[width=0.48\textwidth]{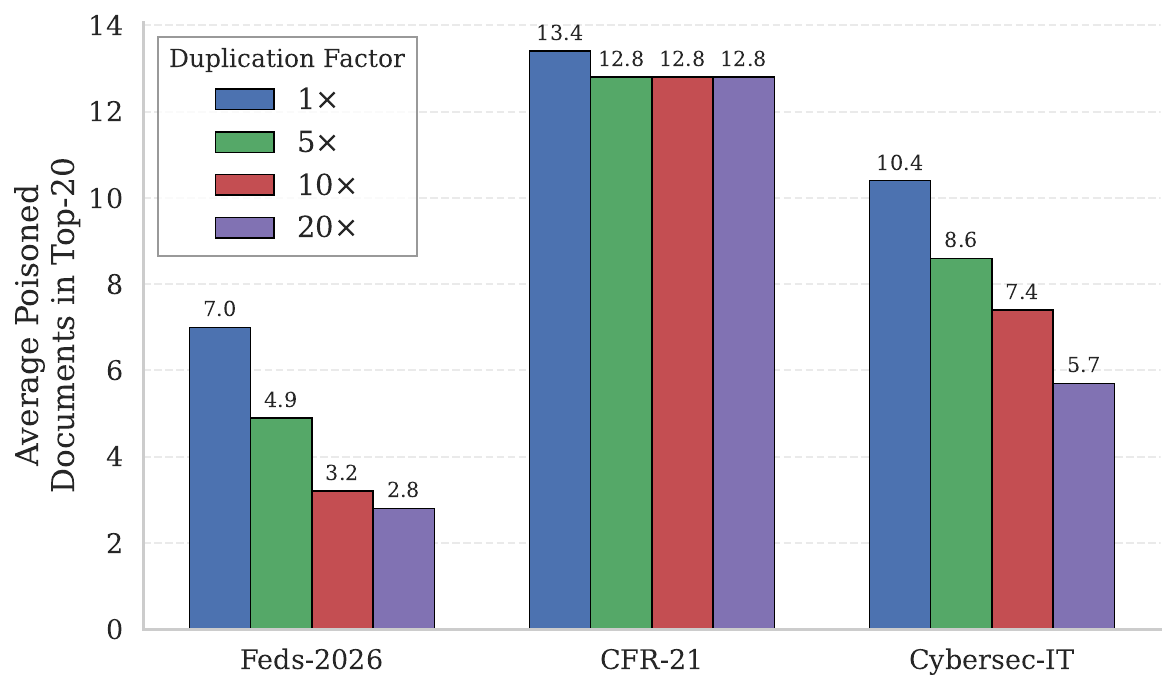}
    \vspace{-15pt}
    \caption{Average number of poisoned documents retrieved in the top-20 results under different benign document duplication factors.}
    \vspace{-15pt}
    \label{fig:dilution_poisoned_docs}
\end{figure}

To evaluate resilience against corpus inflation as a potential defense, we replicate the benign corpus at 5, 10, and 20 times density and measure the mean number of adversarial documents appearing in the top-20 retrieved results. The Figure~\ref{fig:dilution_poisoned_docs} summarizes the results. While increasing corpus density reduces adversarial representation in the retrieval set, \textsc{\name} remains effective across all corpora. CFR-21 exhibits the strongest resilience to dilution, maintaining a high concentration of adversarial documents even under aggressive inflation, whereas Feds-2026 experiences the largest reduction. Nevertheless, adversarial documents consistently persist within the top-20 retrieved results across all settings, indicating that corpus inflation alone is insufficient to prevent injected adversarial content from reaching the generator context.

\vspace{-5pt}

\subsection{Anomaly Detection Evasion}
\label{sec:anomaly}

\subsubsection{Detector Architecture}

We evaluate against a dual-engine anomaly detector representative of defenses proposed for RAG systems~\cite{zhou2025trustrag,song2026adversarialcot}. One Isolation Forest operates on document embeddings, while a second operates on stylometric features. Both are trained on the legitimate corpus, and a document is flagged if either detector identifies it as anomalous.

\subsubsection{Results}

Table~\ref{tab:adversarial_performance} summarizes evasion performance. Across all corpora, \textsc{\name} consistently achieves higher evasion rates than the Naïve Baseline, demonstrating that stylometric mimicry and manifold-anchored synthesis reduce detectability. The largest improvement is observed on Cybersec-IT corpora, suggesting that corpus-specific writing characteristics make stylometric alignment particularly important in technical domains.

\begin{figure}[htbp]
    \vspace{-5pt}
    \centering
    \includegraphics[width=0.5\textwidth]{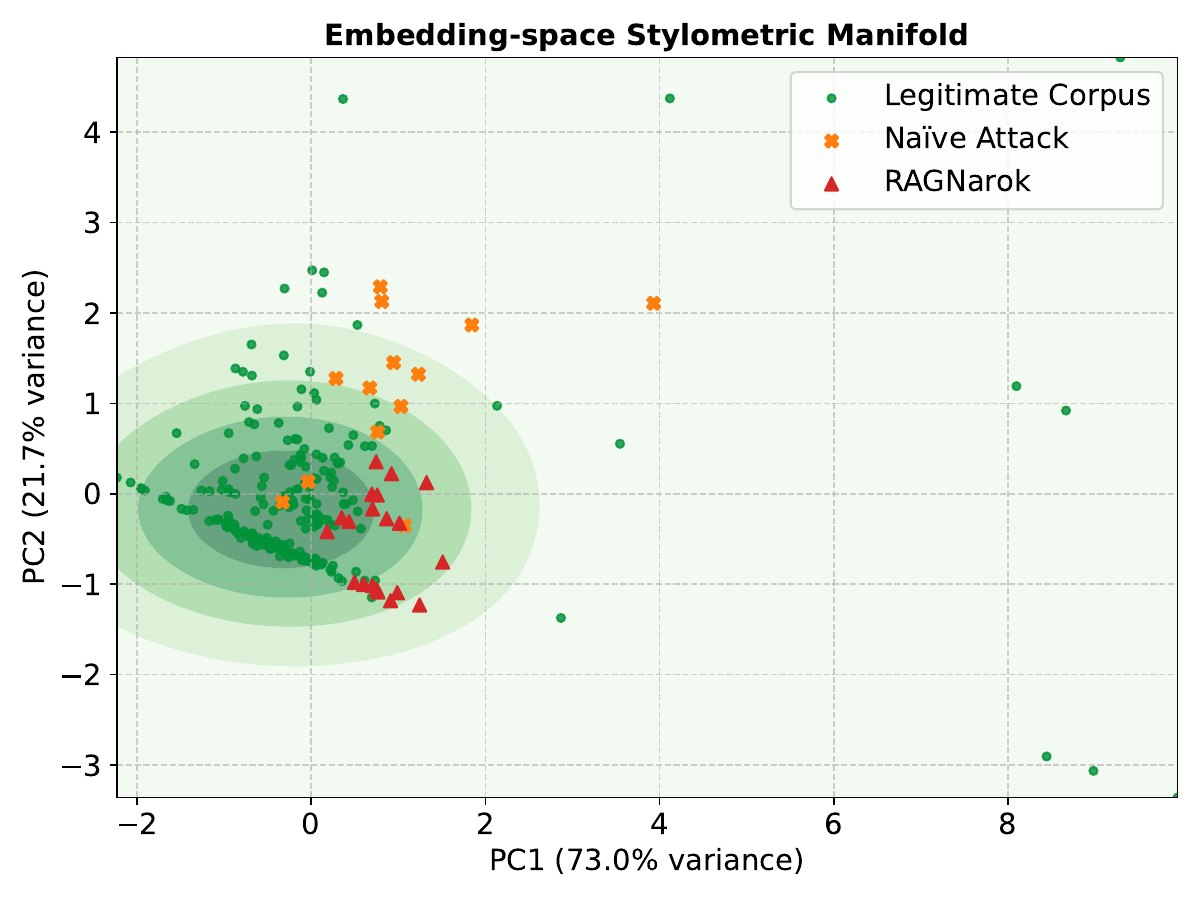}
    \vspace{-25pt}
    \caption{Defense Evasion Analysis.}
    \vspace{-5pt}
    \label{fig:anomaly}
\end{figure}


Figure~\ref{fig:anomaly} provides a geometric view of the stylometric feature space for Feds-2026. Compared to the Naïve Baseline, \textsc{\name}-generated documents cluster substantially closer to legitimate corpus documents, indicating that the stylometric verification loop successfully constrains synthesis toward the benign corpus manifold.

\subsection{Generation Influence}
\label{sec:generation}

\subsubsection{LLM-as-a-Judge}

We use an LLM-as-a-Judge evaluation~\cite{zheng2023judging} to measure whether retrieved adversarial documents alter the semantic conclusions of generated responses. The judge reports (1) a Payload Integration Rate indicating whether the adversarial payload influenced the response and (2) a Mean Cascade Rating (0--10) measuring the extent of that influence.

\subsubsection{Results}

Table~\ref{tab:generation_evaluation} reports generation-stage influence. \textsc{\name} matches or exceeds the Naïve Baseline on Payload Integration Rate across all corpora and consistently achieves higher Mean Cascade Ratings, indicating deeper propagation of adversarial content into generated responses. The largest gain is observed on Feds-2026, mirroring the strong retrieval dominance achieved on that corpus.


These results suggest that retrieval success alone does not fully explain generation influence. The improvements achieved by \textsc{\name} indicate that manifold-anchored construction and authority framing contribute to generation-stage persuasion beyond simply obtaining favorable retrieval ranks.



\begin{table}[t]
    \caption{Adversarial Payload Integration and LLM-as-a-Judge Cascade Ratings.}
    \vspace{-7pt}
    \label{tab:generation_evaluation}
    \resizebox{\columnwidth}{!}{%
        \begin{tabular}{@{}|c|ccc@{}|}
        \toprule
        \textbf{Dataset}             & \textbf{Attack Strategy} & \textbf{Payload Integration Rate} & \textbf{Mean Cascade Rating} \\ \midrule
        \multirow{2}{*}{Cybersec-IT} & Baseline                 & 60\%                              & 6.50                         \\
        \cmidrule(l){2-4}
                                     & \name     & 75\%                              & 7.60                         \\
                                     \midrule
        \multirow{2}{*}{Feds-2026}   & Baseline                 & 70\%                              & 5.57                         \\
        \cmidrule(l){2-4}
                                     & \name     & 80\%                              & 8.00                         \\
                                     \midrule
        \multirow{2}{*}{CFR-21}      & Baseline                 & 80\%                              & 7.25                         \\
        \cmidrule(l){2-4}
                                     & \name     & 80\%                              & 8.00                         \\ \cmidrule(l){1-4} 
        \end{tabular}%
    }
    \vspace{-20pt}
\end{table}

\section{Limitations}
\label{sec:discussion}
\vspace{-2pt}


Several limitations constrain the current work. Firstly, the heuristic-based retriever fingerprinting parameters $(S, R)$ are not learned, and the fingerprinting procedure itself warrants dedicated empirical study, particularly for re-ranking-augmented retrievers. Secondly, LLM-as-a-Judge evaluations are subject to positional and verbosity biases~\cite{zheng2023judging}; structured output mitigates; however, it does not eliminate this concern. Moreover, the threat model assumes the sources are transparent in the RAG pipeline, and would fail under a model where sources cannot be observed. Lastly, the document synthesis and the iterative verification process for the synthesized document is computationally expensive.

\section{Conclusion}
\label{sec:conclusion}
\vspace{-1mm}

We introduced \textsc{\name}, a black-box, context-aware corpus poisoning framework for RAG systems that adapts its adversarial payload to the observed retrieval context before injection. \textsc{\name} consistently ranks high in retrieval, evades anomaly detection in up to 88.37\% of trials, and corrupts the semantic conclusion of generated responses across legal, financial, and cybersecurity corpora. We hope this work motivates a new class of context-aware defenses, including provenance-aware retrieval auditing, adaptive anomaly detection, and generation-layer conflict resolution, that match the sophistication of the threat.

\bibliographystyle{IEEEtran}
\bibliography{References}

\end{document}